\documentclass{article}

     \usepackage[final]{neurips_2020}

\usepackage[utf8]{inputenc} 
\usepackage[T1]{fontenc}    
\usepackage{hyperref}       
\usepackage{url}            
\usepackage{booktabs}       
\usepackage{amsfonts}       
\usepackage{nicefrac}       
\usepackage{microtype}      
\usepackage[utf8]{inputenc} 
\usepackage[T1]{fontenc}    
\usepackage{hyperref}       
\usepackage{url}            
\usepackage{booktabs}       
\usepackage{amsfonts}       
\usepackage{nicefrac}       
\usepackage{microtype}      
\usepackage{amsmath, bm}
\usepackage{amssymb}
\DeclareMathOperator*{\argmax}{argmax}
\usepackage{graphicx}
\usepackage{float}
\usepackage[english]{babel}
\usepackage[utf8]{inputenc}
\usepackage[noend]{algpseudocode}
\usepackage{xcolor}
\usepackage[linesnumbered,ruled,vlined]{algorithm2e}
\setcitestyle{square,numbers}

\usepackage{xcolor}
\title{Quantum Tensor Network in Machine Learning:\\ An Application to Tiny Object Classification}

\author{%
  Fanjie Kong \\
  Department of Electrical and Computer Engineering\\
  Duke University\\
  \texttt{fanjie.kong@duke.edu} \\
  
  \And
  Xiao-Yang Liu \\
  Department of Electrial Engineering\\
  Columbia University \\
  Manhattan, New York, NY !0025 \\
  \texttt{xl2427@columbia.edu} \\
  \AND
  Ricardo Henao \\
  Department of Electrical and Computer Engineering \\
  Duke University \\
  \texttt{ricardo.henao@duke.edu} \\
}

\begin{document}

\maketitle

\begin{abstract}
Tiny object classification problem exists in many machine learning applications like  medical imaging or remote  sensing, where the object of interest usually occupies a small region of the whole image. It is challenging to design an efficient machine learning model with respect to tiny object of interest. Current neural network structures are unable to deal with tiny object efficiently because they are mainly developed for images featured by large scale objects. However, in quantum physics, there is a great theoretical foundation guiding us to analyze the target function for image classification regarding to specific objects size ratio. In our work, we apply Tensor Networks to solve this arising tough machine learning problem. First, we summarize the previous work that connects quantum spin model to image classification and bring the theory into the scenario of tiny object classification. Second, we propose using 2D multi-scale entanglement renormalization ansatz (MERA) to classify tiny objects in image. In the end, our experimental results indicate that tensor network models are effective for tiny object classification problem and potentially will beat state-of-the-art. Our codes will be available online \url{https://github.com/timqqt/MERA_Image_Classification}.
  
\end{abstract}

\section{Introduction}
\label{introduction}
%
Neural networks have achieved the state-of-the-art performance in image classification. However, there are still several tasks where neural network may fall short. Typically, in tiny object classification problem, the task is to classify a very small object in a huge uncorrelated background, where the input has very low signal-to-noise ratio. Recent works showed that with limited size dataset, the convolutional neural networks (CNNs) cannot exceed performance than random classifier on very low signal-to-noise ratio problem. It is very challenging to develop machine learning model regarding to a specific signal-to-noise ratio based on current deep learning theory. However, we see the light in quantum physics. Entanglement entropy bound theory builds a bridge for object ratio and function complexity \cite{eeic}. Motivated by the challenge of tiny object classification, we explore the potential of a tensor network model which can efficiently simulate quantum many-body states in multi-scale.

Recently, there are an increasing number of emerging tensor network applications in machine learning \cite{TNapp1}\cite{TNapp2}. These methods apply Matrix Product State (MPS) to classify MNIST digits. The inputs of these model are flattened 1D tensors, which ignores the spatial correlation of image. And in \cite{GE}, the author also utilized MERA for image classification, which inspires our work. Here, we target at a more interesting and practical valuable problem, namely tiny object classification.


In this work, we propose to use 2D multi-scale entanglement renormalization ansatz (MERA) inspired from \cite{eeic} to match the entanglement entropy of target function by a multi-scale model. Second, we present 3 experiments to illustrate how to successfully train a MERA-structure tensor network model and show the promising results for incorporating quantum physics method into deep learning. Our proposed MERA model can be trained end-to-end under current popular automatic differential framework like TensorFlow and achieves comparable results to current neural network method. 

\section{Problem Statement}
\label{Entanglement Entropy}

We model the image classification problem by a quantum spin model. We then use the entanglement entropy to formulate required function complexity for classification of tiny objects in an image. Finally, we present that MERA can be a promising solution where CNN may fails to perform efficiently. 

\subsection{Problem Formulation for Image Classification by Quantum Spin Model}

Recent works \cite{entanglementarealaw, bb, eeic} indicate that the Hilbert space of target function for image classification problems are equivalent to a quantum spin model.  Considering the set of all functions mapping images to classes $\{0, 1\}$ as $\mathcal{H_I} = \{f:S\rightarrow \mathbb{C}\}$. It is easy to show that the function space $\mathcal{H_I}$ is a Hilbert space with dimension $2^N$. Coincidentally,  a quantum many-body spin system is also dealing with a Hilbert space $\mathcal{H_Q} $ with dimension $2^N$ . In \cite{eeic}, the author pointed out that $\mathcal{H_I}$ and $\mathcal{H_Q}$ are equivalent and isomorphic. It is also easy to connect the quantum spin model to the target function model for image classification. For a $1/2$ spin model, each quantum state has probability for two directions of spins. For the target function of image classification problem, each input pixel contributes probability to the image being one of two classes.

We define a linear transformation $T: \mathcal{H_I}\rightarrow \mathcal{H_Q}$  . If we have a target function $F\in \mathcal{H_I}$  for any image classification problem, we can find this function by looking for a corresponding representation in $\mathcal{H_Q}$,  which is a golden quantum state:
\begin{equation}
|\Psi_G \rangle = \sum_{s\in S} F(s)|s\rangle, \end{equation}
where $s$ is the basis of $\mathcal{H_Q}$  thought of as an image in the set $S$.

Now we have the theoretical foundation which implies that techniques applied on Hilbert space of quantum spin model can also deal with image classification problem. In next section, we will describe how the entanglement entropy formulate the target function $F(s)$ for image classification.

\subsection{Entanglement Entropy of Target Function for Image classification}

In quantum physics, the entanglement entropy describes the degree of quantum states entanglement. Given a quantum system divided into two parts A and B, the entanglement entropy $S_A$ is defined by the reduced density matrix $\rho_A$ for part A: $S_A = - Tr_{A}(\rho_A \log \rho_A)$. Similarly, in an image including an object of interest, it can be divided into two parts A and B too like Fig 1, which gives us a chance to model the relationship of target function complexity and object-to-image ratio for tiny object classification problem. Due to $\mathcal{H_Q} \cong \mathcal{H_I}$ and target functions set $\mathcal{H_T} $, entanglement entropy can be a useful tool to characterize the image classification problem. After we discuss the bound of entanglement entropy, we can have some information about how to construct the model with respect to the object-to-image (O2I) ratio. O2I ratio is defined as the ratio of the object  correlates to the label in the whole image, similar to signal-to-noise ratio. 

Quantum system usually satisfies different scaling of the entanglement, such as area law and volume law. Volume law entanglement is not discussed in this paper, because it is far beyond the problem we want to solve. One can have an example of volume law entangled target function that is the images to be classified are randomly generated in dataset. Area law entangled target function means the pixels only entangled locally, for instance, closed loop recognition task. For general image classification problem, the target function satisfies a sub-volume law. In \cite{eeic} , the author argues that the entanglement entropy of target function for image classification problem is bounded by $S_{AB}\leq rL_{AB}\log 2$, where $r$ is the maximum distance within which pixels entanglement exists and $L_{AB}$ is the length of boundary between region $A$ and $B$ . One can easily incorporate the O2I ratio in this bound. We denote the O2I ratio as $\eta$. For example, most shapes of focused object in the image are close to circle or polygon. Then the entanglement entropy has bound $S_{AB}\sim \mathcal{O}(r\sqrt{\eta})$. For objects closed to a line, the bound of entanglement entropy scales as $S_{AB}\sim \mathcal{O}(r\eta)$. In our problem scenario, we only discuss the first type of object in image. The abstract model of images in object classification problem is shown in Fig 1.

To match the level of the sub-volume law entanglement entropy for target function in image classification problem, we suggest using Multi-scale entanglement renormalization ansatz(MERA), which supports multiple entanglement scales at most $O(S_{AB})\sim L_{AB}$ \cite{multiscale}. 




\begin{figure}[t]
\centering
\includegraphics[scale = 0.8]{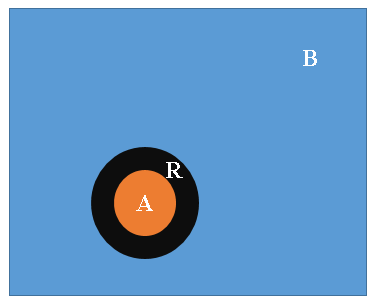}
\label{fig:birds}
\caption{Representation of a general image. Region B is the background. Region A is an object of interest. Region R is the boundary between B and A. }
\label{nn}

\end{figure}


\section{Multi-scale Entanglement Renormalization Ansatz for Image Classification}

In this section, we describe how we define the 2D MERA model and how we achieve optimization. A traditional MERA is a tensor network that corresponds to quantum circuit and the model is finally contracted to 1. Therefore, we make some modifications to the original model and 

The MERA model is basically a cascade of isometries and disentanglers. We follow \cite{MERA}'s work to construct our 2D MERA for machine learning. We define the isometries and disentanglers as follow: 

(i)   tensors in odd rows are disentanglers:
\begin{equation}
 \sum_{\mu_1 \mu_2 \mu_3 \mu_4} (u^{\star})^{\alpha_1 \alpha_2 \alpha_3 \alpha_4}_{\mu_1 \mu_2 \mu_3 \mu_4} (u)^{\alpha_1' \alpha_2 '\alpha_3' \alpha_4'}_{\mu_1 \mu_2 \mu_3 \mu_4}=\delta_{\alpha_1 \alpha_1'} \delta_{\alpha_2 \alpha_2 '} \delta_{\alpha_3 \alpha_3'} \delta_{\alpha_4 \alpha_4'}
\end{equation}

(ii)   tensors in even rows are isometries:
\begin{equation} 
\sum_{\mu_1 \mu_2 \mu_3 \mu_4} (w^{\star})^{\alpha}_{\mu_1 \mu_2 \mu_3 \mu_4} (w)^{\alpha}_{\mu_1 \mu_2 \mu_3 \mu_4}=\delta_{\alpha \alpha'} 
\end{equation}

(iii) Also, we assign output indices for the top tensor:
\begin{equation}
\sum_{\mu_1 \mu_2 \mu_3 \mu_4} (t^{\star})^{\alpha}_{\mu_1 \mu_2 \mu_3 \mu_4} (t)^{\alpha}_{\mu_1 \mu_2 \mu_3 \mu_4}=\delta_{\alpha \alpha'} 
\end{equation}
which is in the same number of classes for image classification.\cite{MERA} And for the input, we let each pixel of the input image as the bottom tensors. We divide the whole image into $4x4$ regions and pixels in the same region are connected to the same disentangler, which is shown in Fig 2. 
In this modified version of MERA, we then can train this model on our tiny object classification problem.

The optimization method we applied is the same as \cite{TNML}. Here, we use automatic differentiation built in TensorFlow to minimize a cross entropy loss function of labels and output from 2D MERA. More precisely, our loss function is 
\begin{equation} L = -\sum_{(x_i, y_i)\in \mathcal{D}}y_i \log \text{softmax}(\delta(x_i)).
\end{equation}
And the prediction is 
\begin{equation}
f(x) = \argmax_{c} \text{softmax}(\delta^{(c)}(x_i))
\end{equation}

Under this configuration, we can easily train the 2D MERA model using TensorNetwork with TensorFlow backend. Similar to training a neural network, in training step, the input of the algorithm is a pair of image and label. The input image is divided into $N^2$ regions. Each region is flattened into a one dimensional tensor at the bottom of the MERA. Then, the bottom tensors are contracted with disentanglers and isometries alternatively, until the contraction reaches the top tensor. The loss function takes the top tensor and the label of the image. Gradient descent is used to minimize the loss function and updates all parameters in disentanglers and isometries by back-propagation. In inference step, the predication of classification is the argument of the maxima for the top tensor.

Though the brute-force gradient method has some flaws \cite{TNML}, we show that it still can successfully optimize the model to a reasonable performance.

\section{Performance Evaluation}

We demonstrate that the tensor networks can be a promising tool to tackle the tiny object classification problem where CNN may fail. We evaluate the performance of the MERA algorithm on MNIST, NeedleMNIST and LIDC datasets, and compare with conventional neural networks.

\begin{figure}[t]
\centering
\includegraphics[scale=0.5]{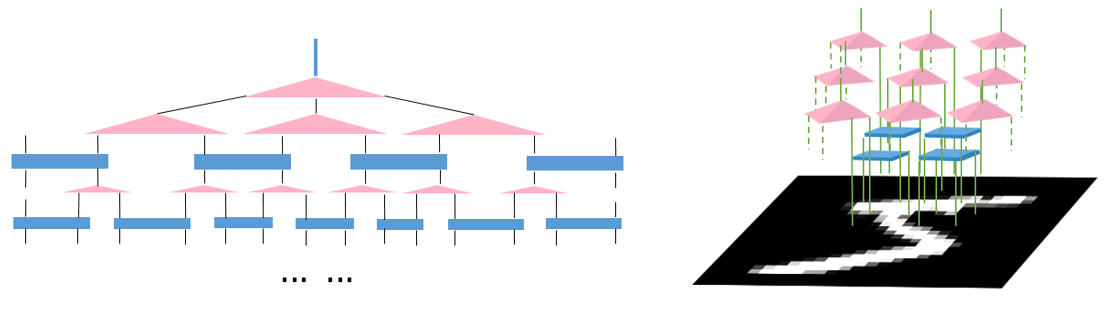}
\caption{MERA in 1D (left) and 2D (right). Inspired by \cite{GE} and redrawn it.}
\label{data}
\end{figure}

\subsection{Data Sets}

Our experiments are performed on three public datasets, where we explore hyper-parameter tunning on MNIST dataset and NeedleMNIST and LIDC datasets are associated with our focused task of tiny object classification. 

\textbf{MNIST: } MNIST dataset is widely used for judging the performance of heuristic machine learning model, including a large collection of handwritten digits. Images in the dataset have $28x28$ pxiels in grey level and correspond to a class label from $0$ to $9$ \cite{MNIST}. In our experiments, MNIST dataset serves as a playground for understanding hyper-parameter tuning and demonstrates the powerfulness of tensor network techniques. 

\textbf{NeedleMNIST: } NeedleMNIST dataset is introduced in \cite{needle}, which is inspired by the cluttered MNIST dataset \cite{cluttered}. In NeedleMNISt dataset, each image is controlled at a specific object-to-image ratio by randomly pasting a MNIST digit patch on the black background. Also, some digit patches unrelated to the classification label are also added as input noise. The dataset includes images with O2I ratio range from 19.1\% to 0.075\% with corresponding size from $64\times 64$ to $1024\times 1024$ . Here, we choose digit 3 as the object of interest to match the experimental setting in \cite{needle}. Due to the computational limit, we only run our experiments on dataset with size $64\times 64$ and $128\times 128$. The NeedleMNIST dataset is a great testbed for evaluating the performance of tensor network models in tiny object classification problems.
\\
\textbf{LIDC: }The Lung Image Database Consortium image collection (LIDC) is a dataset including clinical thoracic CT scans with annotated lesion area. And there are 1018 CT images and associated XML file for lesion masks. In our experiment, we take advantage of this dataset for classification task, similar to \cite{TNmedical}. If one CT image has an associated lesion mask, we label it as positive sample, and vice versa. The lesion area has an average O2I ratio 1.32\% \cite{LIDC}, which is a very typical tiny oject problem. 


\subsection{Experimental Setting and Results}
In our experiments, we compare the performance of three machine learning models. And also, we explore how the hyper-parameters influence the performance of 2D MERA. 

In our experiments on MNIST dataset, we aim to explores the critical hyper-parameters to ensure the convergence of the model. During our experiments, we find that the learning rate of back-propagation optimizer and the standard deviation of initial model parameters are essentially important to model convergence. In Fig 6 and Fig 7, we plot the loss and accuracy during training with different learning rates and standard deviations. In the end, we decide the best parameter combination is \{lr: 1e-5, std: 0.0001\}. 

After we tuned the model on MNIST dataset, we then train our model on NeedleMNIST dataset with size 64x64 and LIDC dataset with size 128x128. In these experiments, the 2D MERA model inherits the same hyper-parameters from MNIST experiments. Each input tensor(bottom tensor) represents one pixel in the image and has only one dimension. 

To compare our 2D MERA model with neural network baseline, we use a Alex-Net \cite{AlexNet} like neural network for MNIST experiment and for NeedleMNIST and LIDC, DenseNet\cite{DenseNet} are used in these experiments as comparison, similar to \cite{TNmedical}. 

Also, we include the performance of neural network model that has a tensor network layer in the experimental results. The tensor network module here is locally orderless fusing with globally orderless inputs \cite{TNmedical}, which yields better results when it replaces a convolutional layer. The aim of this experiments is to show the tensor network not only has potential to beat neural network but also can boost the performance of current state-of-the-art.


\begin{table}[t]
\caption{Experimental results on MNIST, NeedleMNIST and LIDC datasets.} 
\centering 
\begin{tabular}{c c c c c} 
\hline\hline 
 & MNIST  & NeedleMNIST ($64\times 64$) &  NeedleMNIST ($128\times 128$)  & LIDC  \\ [0.5ex] 
\hline 
 CNN  & $98.3 \%$   & $76.0\%$ & $73.9\%$   & $78.0\%$ \\ 
 Tensor-NN & $98.5\%$   & $74.0\%$ &$72.7 \%$ & $86.0\%$ \\
2D MERA & $90.3\%$   & $78.4\%$ & $71.4\%$  & $76.0\%$ \\
[1ex] 
\hline 
\end{tabular}
\label{table:nonlin} 
\end{table}

         
\begin{figure}[ht]
\centering
\includegraphics[scale=0.6]{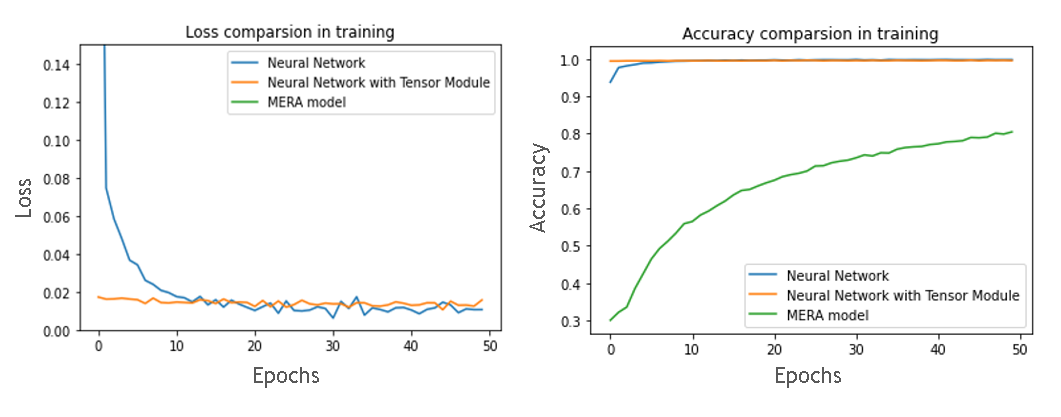}
\caption{Evolution of training comparison. }
\label{comparing}
\end{figure}
\small

\begin{figure}[H]
\centering
\includegraphics[scale=0.6]{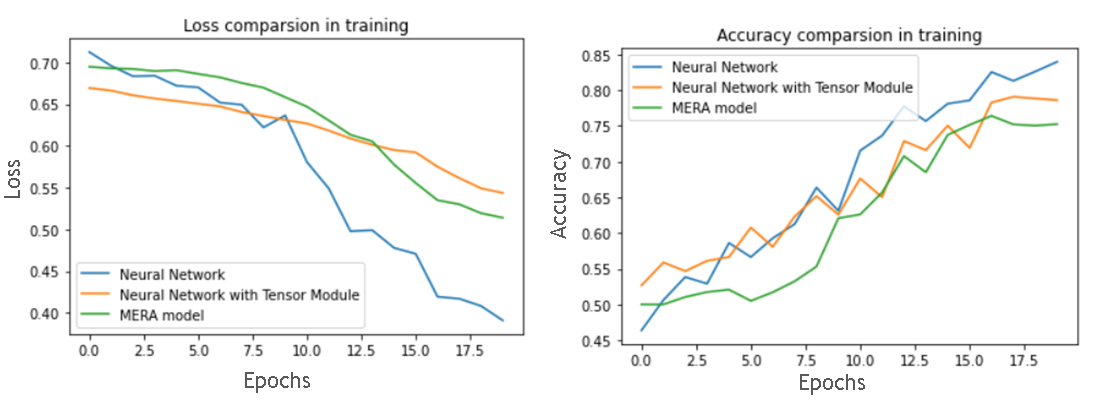}
\caption{Loss and accuracy of MERA during training on NeedleMNIST dataset with size $64\times 64$. }
\label{comparing}
\end{figure}
\small


\begin{figure}[H]
\centering
\includegraphics[scale=0.6]{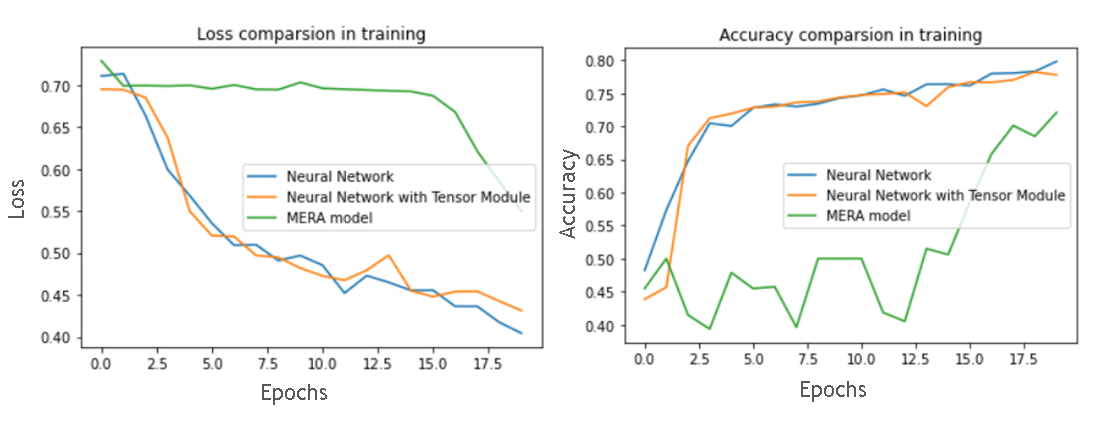}
\caption{Loss and accuracy of MERA during training on NeedleMNIST dataset with size $128\times 128$. }
\label{comparing}
\end{figure}


The performance of three models for MNIST, NeedleMNIST and LIDC datasets are reported above. We show that 2D MERA yields comparable performance to the neural network. And the neural network with tensor network layer model exceeds the traditional neural network in MNIST and LIDC datasets. These results imply that the tensor network are potentially useful to boost the neural network or even take its crown in deep learning field. 

We also plotted training loss and accuracy curves for MNIST dataset. The goal is to show that the learning rate of optimizer and standard deviation of initial parameters play important role to yield a converged MERA model. Current gradient descent method to optimize MERA model may be not very ideal for efficient training. Efficiently using tensor network model is still an open question.

\begin{figure}[ht]
\centering
\includegraphics[scale=0.6]{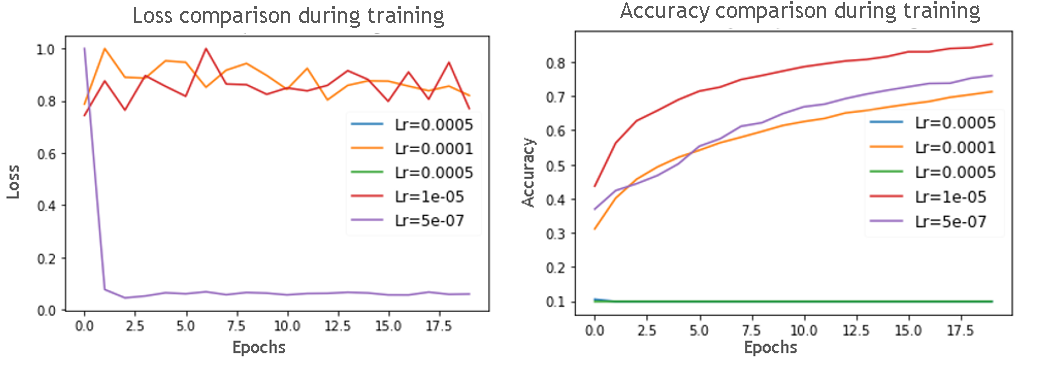}
\caption{Loss and accuracy of MERA during training on MNIST dataset. }
\label{comparing}
\end{figure}
\begin{figure}[ht]
\centering
\includegraphics[scale=0.6]{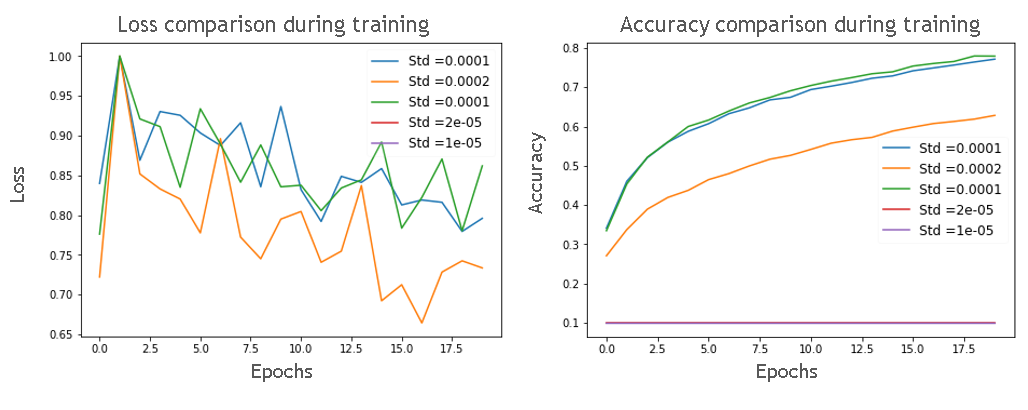}
\caption{Loss and accuracy of MERA during training on MNIST dataset, with different standard deviation for parameters initialization. }
\label{comparing}
\end{figure}

\small
\small
\section{Conclusion}
In this paper, we have summarized the theory in previous work to explain why tensor networks is a promising solution to tiny object classification problem. Second, we propose a 2D MERA model optimized by gradient. Third, we compare tensor network models with neural network on MNIST and tiny object dataset including NeedleMNIST and LIDC. We also explore the influence of hyper-parameters for tensor network model. Based on the comparison, we show the tensor network is a promising tool to tackle tiny object classification problem. Future work will be interesting to explore more optimization techniques for tensor network\cite{GE}, tricks to improve performance\cite{TNmedical} and accelerating training and inference time\cite{accTN}.


\end{document}